\pdfoutput=1

\documentclass[11pt]{article}

\usepackage[]{acl}

\usepackage{times}
\usepackage{latexsym}

\usepackage[T1]{fontenc}

\usepackage[utf8]{inputenc}
\usepackage{textcomp}

\usepackage{times}
\usepackage{latexsym}
\usepackage{hyperref}
\usepackage{graphicx}
\usepackage{algorithm}
\usepackage{algpseudocode}
\usepackage{amsfonts}
\usepackage{listings}
\usepackage{mathtools}
\usepackage{booktabs}
\usepackage{multirow}
\usepackage{xcolor}
\usepackage{graphicx}
\usepackage{subfigure}
\usepackage{float}
\usepackage{lipsum}
\usepackage{stfloats}
\usepackage{authblk}

\usepackage{inconsolata}
\usepackage{textcomp}

\title{GenSco: Can Question Decomposition based Passage Alignment improve  Question Answering?}

\author{
Barah Fazili$^1$\thanks{Research work conducted during internship at Adobe Research India. }, Koustava Goswami$^2$, Natwar Modani$^2$, {Inderjeet Nair}$^3$ \thanks{Research work conducted when at Adobe Research India.} \\
$^1$ Indian Institute of Technology, Bombay \\
$^2$ Adobe Research, India\\
$^3$ University of Michigan, Ann Arbor\\
{\tt \small barah@cse.iitb.ac.in},
{\tt \small \{koustavag,nmodani\}@adobe.com},
{\tt \small nderjeetnair1@gmail.com}
}

\begin{document}
\maketitle
\begin{abstract}
Retrieval augmented generation (RAG) with large language models (LLMs) for Question Answering (QA) entails furnishing relevant context within the prompt to facilitate the LLM in answer generation. During the generation, inaccuracies or hallucinations frequently occur due to two primary factors: inadequate or distracting context in the prompts, and the inability of LLMs to effectively reason through the facts. In this paper, we investigate whether providing \textit{aligned} context via a carefully selected passage sequence leads to better answer generation by the LLM for multi-hop QA. We introduce, \textit{``GenSco'', a novel approach of selecting passages based on the predicted decomposition of the multi-hop questions}. The framework consists of two distinct LLMs: \textbf{(i)} \textit{Generator} LLM, which is used for question decomposition and final answer generation; \textbf{(ii)} an auxiliary open-sourced LLM, used as the \textit{scorer}, to semantically guide the \textit{Generator} for passage selection. The generator is invoked only once for the answer generation, resulting in a cost-effective and efficient approach.\footnote{For question decomposition,the generator is called $O(N)$ times where $N$ is the max number of hops along the greedy path. Refer to Section~\ref{methodology} for details.} We evaluate on three broadly established multi-hop question answering datasets: 2WikiMultiHop, Adversarial HotPotQA and MuSiQue and achieve an absolute gain of \textbf{$15.1$} and \textbf{$5.9$} points in Exact Match score with respect to the best performing baselines over MuSiQue and 2WikiMultiHop respectively. 
\end{abstract}

\section{Introduction}

Retrieval augmented generation (RAG) with Question answering typically involves presenting the model with a ``context" supporting the ground truth answer, such as a paragraph from wikipedia, alongside the posed question. This task offers a measurable means to assess the language comprehension and reasoning capabilities of an NLP system~\cite{hermann2015teaching,xiao2019dynamically,rajpurkar2016squad}. Earlier approaches predominantly concentrated on conducting this reasoning within a singular context~\cite{liu2018stochastic, seo2018bidirectional, wang-etal-2017-gated}. Thanks to recent advancements in Deep Learning techniques~\cite{lan2020albert}, machines have now surpassed human performance on datasets like SQUAD 2.0~\cite{rajpurkar2016squad}. The recent progress in single-hop QA tasks has spurred interest towards more challenging and practical QA form, Multi-Hop Question Answering (MHQA). Multi-step reasoning involves asking one or more preliminary questions before getting to the final answer, with each preliminary step feeding into the subsequent step, forming a reasoning chain~\cite{mavi2022survey}. Consequently, techniques proven effective in MHQA can be seamlessly integrated into tasks such as sentence fusion~\cite{geva2019discofuse,weiss2021extending}, abstractive summarization~\cite{nayeem-etal-2018-abstractive}, event occurrence time prediction~\cite{10.1145/3404835.3462885}, multi-document summarization~\cite{ma2021multidocument}, and timeline summarization~\cite{yu-etal-2021-multi}, all of which require information synthesis across multiple documents.

While LLMs produce efficient results for most of the natural language understanding tasks, finding the best way to provide intructions to LLMs to perform MHQA remains a popular area of research. 

Factual inaccuracy occurs when the model lacks the requisite supporting data to generate an accurate response, often arising from its unfamiliarity with specific entities, attributes, or events. Although this kind of inaccuracy is simple, it constitutes the bulk of errors in the model generations~\cite{zheng2023does}. 
For multi-hop QA, there has been some work on language model prompting for multi-hop passage re-ranking where the passage relevance scores are computed using conditional likelihoods using the LLM~\cite{khalifa2023fewshot}. Another thread of research focuses on providing simplified queries to LLMs by performing  Question Decomposition~\cite{patel2022question}. Question decomposition has been explored by either involving human in the loop ~\cite{patel2022question} or getting the model to respond to subquestions before rendering the final answer~\cite{radhakrishnan2023question,schlag2023large,yao2023tree}. But generating answers at every iteration for each of the decomposed questions for longer sequence of documents is time consuming and a tedious process to solve. Infact there is a high risk of hallucination if the passage selection goes wrong.

To tackle this problem, we explore \textit{Question Decomposition} for \textit{passage retrieval} instead for invoking the LLM for answer generation at each step, resulting in less latency and higher efficiency. 

We combine the two kinds of approaches: the ones leveraging instruction tuned LLMs to compute relevance scores in passage reranking and the methods generating simpler questions via Question Decomposition, to design an effective passage sequence selection method for MHQA. We propose \textit{GenSco} which leverages an open-source LLM as a scorer guiding the Generator LLM (assumed to be a black box) for passage sequence selection before Answer Generation. Our intuition behind choosing two separate LLMs is to complement the generator LLM with a scorer module in terms of semantic and grounded knowledge base.

In our proposed approach, we start with an empty context. We ask generator LLM to generate a sub-question from the question, the context collected up to now (initially empty), as well as, the sub-questions generated up to now (again, initially empty). Given the generated sub-question, we rank all the candidate passages based on negative log-likelihood using the scorer LLM. We add the passage with the best score to the context (and the generated sub-question to the list of sub-questions) and ask the generator LLM to generate next sub-question. When the stopping criteria is met, we send the context (in the order it was accumulated) and the question as part of a suitable prompt to the generator LLM for final answer generation.

To the best of our knowledge, we are the first to propose Question Decomposition for selecting a passage sequence for multi-hop question answering. Note that it’s different from simple reranking since there are two aspects to passage selection with our method: not only do we identify the most relevant passages but also render them in an order that is consistent with the reasoning steps implied by the multi-hop question. We empirically show that GenSco retrieves relevant passages with a high precision and that the order in which the passages are presented to the downstream LLM also contributes to achieving higher accuracy. To summarize, our main contributions in this paper are: \textbf{(i)}: We introduce a novel, inference only (hence data-efficient) greedy approach called \textit{``GenSco''} for passage sequence selection in Multi-Hop Question Answering and achieve an absolute gain of $15.1$ and $5.9$ points in Exact Match score wrt best performing baselines for 2WikiMultiHop~\cite{ho2020constructing} and MuSiQue~\cite{trivedi-etal-2022-musique} datasets respectively.

; \textbf{(ii)} Apart from the superior downstream QA performance, \textit{``GenSco''} also achieves high precision on the passage retrieval task, effectively mitigating hallucination in the LLM responses.

\section{Related Work}
\subsection{LLMs for Information Retrieval}

Researchers have experimented with in-domain few-shot examples with LLMs to generate queries~\cite{dai2022promptagator,bonifacio2022inpars,boytsov2023inparslight}

Thereafter, neural retrieval models are fine-tuned over this enhanced dataset. 

While LLMs have been previously used to score paragraphs based on their relevance to a certain query~\cite{sachan-etal-2022-improving}, directly applying them to determine pertinent paragraphs based on a complex query requiring intricate reasoning results in sub-optimal performance. Our novel approach of breaking down a complex query into a set of simpler elemental queries allows our method to leverage generative ability of LLMs for accurate retrieval and better answering performance. \cite{zhang2023beam} introduce beam retrieval for multihop QA optimizing learnable parameters across all hops. In contrast, our suggested method exclusively employs an inference approach, eliminating the necessity for training or training examples.

\subsection{Complex Task Decomposition}
One of the well known techniques for MHQA is to decompose the complex query into simpler sub-questions, answer them and then combine the results to get the overall answer~\cite{fu2021decomposing}. Recently researchers have proposed deconstruct a complex problem into a series of simpler sub-problems before feeding to LLMs~\cite{zhou2022least,press2022measuring}. 
 
Despite significant advancements in LLMs that have enhanced their reasoning capabilities and reduced the disparity between machine and human intelligence, using them directly to answer sub-questions might result in inaccurate responses due to the lack of appropriate knowledge. Thus, instead of directly solving the sub-problems via prompting, we leverage LLMs in finding the right context for each of the sub-problems. This is inline with the conclusions drawn by ~\citet{zheng2023does} - providing fine-grained external knowledge can boost the truthfulness of LLMs in answering a question.

\section{Methodology}
\label{methodology}
\begin{figure*}[t!]
\small
   \centering    \includegraphics[width=\textwidth, height=8cm]{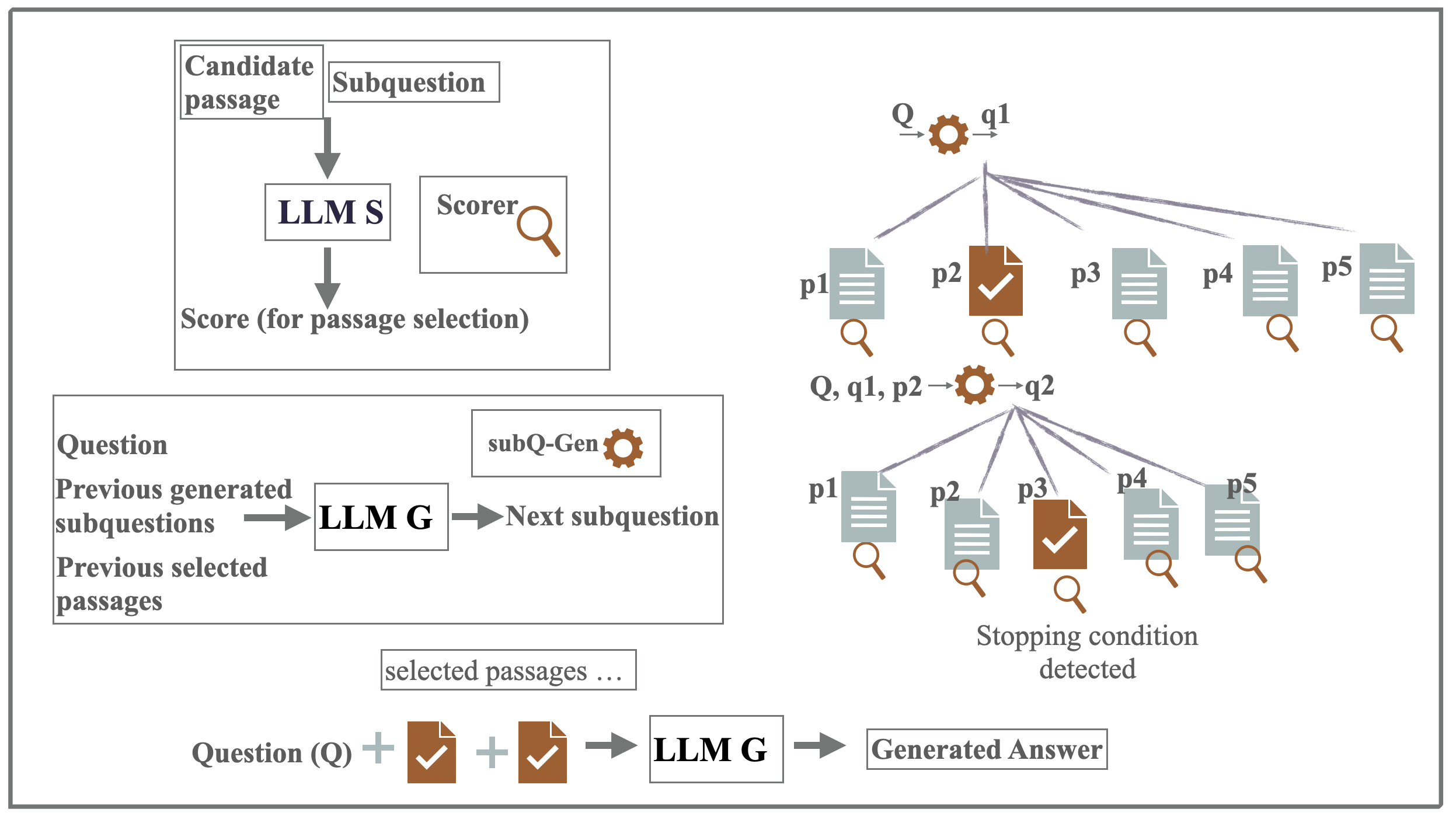} 
\caption{GenSco : subquestion at each level is generated using subQ-Gen module , the Scorer module is invoked for selecting the passage (greedy algorithm). The sequence of passages are then passed as context to G to generate the final answer (bottom) }
   \label{fig:algo}
\end{figure*}

We introduce ``GenSco" for passage selection in retrieval augmented Multi-Hop Question Answering (refer to Figure~\ref{fig:algo}). GenSco leverages two different LLMs (\textit{scorer S} and \textit{generator G}) in a greedy exploration of a passage tree. In order to improve the alignment of selected passages with the reasoning chain implied from the multi-hop question, GenSco provides a refined passage selection technique by using the \textit{generator} for question decomposition in tandem with the \textit{scorer} for passage selection at each step. 

For a given question $Q$ and a set of already retrieved passages $P = [p_1,p_2,...,p_k]$, we construct a tree of passages where each node along the greedy path is expanded to k child nodes $n_{i,1},...n_{i,k}$\footnote{The first element in the subscript indicates the level of the node in the tree(0-indexed, starting from the root node) and the next element represents the index of the passage included at this node.}  each corresponding to one of the $k$ passages in $P$. Node $n_{i,j}$ represents the history of passages included on the path from the root to its parent node along with the passage $p_j$ included at the $i^{th}$ level. 

The subquestion $q_i$ for each level $i$ in the tree is drawn out of the generator LLM \textit{G} using few-shot prompting. More detail on the prompt structure is available in the supplementary.
The nodes at level $i$ are evaluated using \textit{scorer S} for relevance  to the subquestion $q_i$ which presumably captures the $i^{th}$ reasoning step for the multi-hop question $Q$. The node with the best score is chosen from among the $k$ candidates at each level, which is then used to create the next batch of candidates up to level of the $last$ subquestion. Since the subquestions could be indefinitely generated (using \textit{G}), note that it requires specifying an upper limit on the number of levels that can be explored, hence we introduce a novel automated stopping criterion leveraging the \textit{scorer} model in addition to setting a rough upper bound estimated for the dataset. A broad outline is presented in Algorithm~\ref{code} while we expand on more details in section {\color{red}~\ref{sec:question_decomposition}} and section {\color{red}~\ref{sec:passage_selection}} which serve as the two primary modules of GenSco technique.
\begin{algorithm}
\caption{GenSco} 
\small
\label{code}
\begin{algorithmic}[1]
\State G: Generator LLM, S: Scorer LLM
\For {(P,Q,A) in data}
    \State $P : [p_{1},p_{2},...p_{k}]$
    \For {i in 1 to max-levels}
        \If {i = 1}
            \State $q[i] \gets G(Q,shots')$
        \Else
            \State $q[i] \gets G(Q,shots',q[1:i-1],P'[1:i-1])$
        \EndIf
        \If {stop}
            \State break
        \EndIf
        \State $P'[i] \gets S(q[i],P)$
    \EndFor
\State $A' \gets G(Q,P',shots)$
\State Compute\_Metrics$(A',A,P)$
 \State \EndFor
\end{algorithmic} 
\end{algorithm}

\subsection{Question Decomposition} \label{sec:question_decomposition}
For decomposition, the multi-hop Question $Q$ is provided to \textit{G} along with the previously generated subquestions and corresponding passages selected for each subquestion. The prompt also includes an instruction to flag the end of decomposition when no more subquestions can be created for $Q$ by generating a specified keyword. Apart from relying on \textit{G} to flag the end of decomposition or detecting a repeated subquestion, we propose a stopping criterion based on the log-likelihood scores from \textit{S}:
Scorer model \textit{S} is instructed to generate a multihop question given the decomposition and we evaluate the following expression where $nll_S$ refers to negative log-likelihood from Scorer \textit{S}:\\ 
\begin{equation*}
\label{stop-exp}
\begin{multlined}
    nll_S(p(Q|C,q_1,q_2,...q_i)) > \\ nll_S(p(Q|C,q_1,q_2,...q_{i-1})) \\
    C = P'[j], j \epsilon [1,i-1] \tag{1}
\end{multlined}
\end{equation*}

When the likelihood given the decomposition  including the generated subquestion $q_i$ falls below the likelihood while excluding $q_i$ from the prompt, we stop exploring more passages.
Note that the prompt to \textit{S} also includes $C$ which represents the concatenation of passages selected for $q_1,...,q_{i-1}$ represented as $:P'_1,...,P'_{i-1}$. This is included to provide the model enough context to synthesize(indirectly score) the composite question $Q$. The prompt is designed only to extract likelihood scores from the scorer \textit{S} in order to get a proxy for the significance of the new subquestion $q_i$ in the decomposition of $Q$, therefore, take note that we ignore the question generated here by \textit{S}.
\subsection{Passage selection} \label{sec:passage_selection}
Once the subquestion $q_m$ has been formulated for level $m$, we proceed to expand the most favorable node from the previous level $(m-1)$ into k child nodes. Each of these child nodes, denoted as $n_{m,i}$, represents the concatenated sequence of passages chosen along the greedy path from the root node to the parent of $n_{m,i}$, followed by the passage $p_i$. For each child node at level $m$, a score is computed relative to the subquestion $q_m$ using the scorer LLM \textit{S}. The LLM \textit{S} is instructed to produce a question based on the passage sequence represented by each node at level $m$, yielding $m$ distinct scores for subquestion $q_m$ being posed as the output text sequence.

The score for node $n_{m,i}$ is computed by taking the log likelihood of the token sequence $q_m$ in the output distribution of \textit{S} given the already selected passages($P'[:m-1]$) plus passage $p_i$ to the scorer \textit{S}.
\begin{equation*}
\begin{multlined}
    \label{eq:score}
 score(n_{m,i}) =  nll_S(p(q_m/P'[:m-1],P[i])), \\ i\epsilon[1,k] \tag{2}
\end{multlined}
\end{equation*}
After the passage selection is done, the generator is called to which we provide the multi-hop question and selected passage sequence in a few-shot setting to generate the answer.

\begin{equation*}
\Tilde{A} =  G(Q,P') \tag{3}
\end{equation*}
Here $P'$ denotes the sequence of passages chosen following a greedy approach, with the final level being either the maximum permitted levels or determined by stopping conditions, whichever is reached first.\\

Note that in our log-likelihood expressions, we compute the probability of the question conditioned on the passages rather than the other way around which may be less intuitive since we should be scoring the passages(and not the question). As has been discussed in prior work~\cite{sachan2023improving}, either of the two forms could be used theoretically (using the Baye's rule and taking passage priors as uniform which is a reasonable assumption for reranking). This form is more practical(and faster) since passages tend to be much longer than the question and the likelihood of longer sequences could approach zero.

\section{Experiments}
\subsection{Datasets}
We have conducted experiments on three different datasets: \textbf{(i)} 2WikiMultiHop~\cite{ho2020constructing}, \textbf{(ii)} Adversarial HotPot~\cite{yang-etal-2018-hotpotqa,ye2022the}; \textbf{(iii)} MuSiQue~\cite{trivedi-etal-2022-musique}. More details of the datasets can be found in Section \ref{app:dataset} in Appendix.

\subsection{Baselines}

We start with retrieval methods, BM25{~\cite{robertson1994some}} and GTR~\cite{ni2021large}, by extracting the most relevant passages, and fed to the LLM for answer generation. 
We compare against Verify-and-Edit~\cite{zhao2023VerifyandEdit} (CoT-SC + VE), that focuses on post-editing `Chain of Thought'-style reasoning, 
`Iter-Retgen'~\cite{shao2023enhancing}, ReAct~\cite{yao2023react} and SelfAsk~\cite{press2023measuring}. 
Both ReAct ~\cite{yao2023react}  and Self-Ask~\cite{press2023measuring} are approaches that involve the decomposition of complex/multihop questions using Large Language Model (LLM) prompting techniques.Also, we compare with a cross-encoder\footnote{\href{https://huggingface.co/cross-encoder/ms-marco-MiniLM-L-6-v2}{https://huggingface.co/cross-encoder/ms-marco-MiniLM-L-6-v2}} trained for MS Marco Passage Ranking. Our apporach of GenSco is an inference method, hence
not directly comparable to the full finetuning setup. We draw inspiration from different approaches researchers have tried to make zero-shot/few-shot settings effective for the MHQA~\cite{gao2024retrievalaugmented}. The detailed explanation can be found in Apendix.
\subsection{Experiment Details}
For 2WikiMultiHop, as the dataset has atmost 5 supporting passages per instance, we select the top-5 passages for each of the \textit{implemented baselines} (top three rows against each dataset in Table~\ref{main_table}). The passages are then provided as context in the prompt to the LLM for QA. Also, as part of the prompt we provide two instances from the train set for our few-shot prompting of the Generator LLM . For Adversarial HotPot, we provide top-2 passages using each of the three implemented baseline methods, prompting the Generator LLM in the same way but with 4-shots. For MuSiQue, we augment the prompt with the top-4 passages for each of the three implemented baselines and include 3 shots of train instances in the prompt.

For 2WikiMultiHop, \textit{GenSco} explores  the search tree to a limit of 5 levels, considering that each instance has a maximum of 5 supporting passages. In terms of prompting the Generator LLM  for GenSco , we include two instances from the training set as for the baseline retrieval methods. For the Adversarial HotPot dataset and MuSiQue similarly, we utilize 4 shots and 3 shots of training instances respectively as for the implemented baselines. \textit{Verify-and-Edit}~\cite{zhao2023VerifyandEdit} however, uses six shots for both its datasets.\\
We leverage GPT3.5\footnote{\href{https://platform.openai.com/docs/models/gpt-3-5}{text-davinci-003}} model as the generator LLM for all techniques as has been used for the \textit{Verify-and-Edit} baseline. For GenSco,  3B version of Open-llama\footnote{\href{https://huggingface.co/openlm-research/open_llama_3b}{Open LLAMA on Hugging Face }} is taken as the scorer LLM. 
We implement two main variations of our GenSco approach. The first variant, named GenSco-stop, includes the scorer log-likelihood based condition in inequality (\ref{stop-exp}) as an additional stopping criteria. The second variant, called GenSco-max, either detects the specified end-of-decomposition keyword in the model output or identifies redundant subquestions (already generated for the instance) as the stopping conditions. Note that GenSco-stop incorporates the likelihood based stopping criterion as a third alternative beyond the two criteria in GenSco-max to stop the greedy search.\\

The scores of CoT-SC+VE~\cite{zhao2023VerifyandEdit}, ReAct~\cite{yao2023react}, Self-Ask~\cite{press2023measuring} and Iter-Retgen~\cite{shao2023enhancing} for all three datasets have been directly taken from prior work~\cite{zhao2023VerifyandEdit,shao2023enhancing} while all other methods listed in the tables are locally evaluated. 
\subsubsection{Metrics}
The predicted answers are evaluated for correctness given the ground truth answers in the datasets. We evaluate Exact Match, F1, Precision and Recall based on the implementation in prior work~\cite{zhao-etal-2023-verify} which simply normalizes the two strings by removing punctuation, lower-casing the text, etc., before matching them word by word for Exact Match (EM) binary score for each instance. Precision computes the fraction of words in the predicted answer overlapping with the ground truth answer. Similarly, Recall measures the fraction of words in the ground truth matching the predicted answer. F1 score is computed in the standard way based on the precision and recall values. 

\subsection{Results}

\begin{table*}
\centering
\small

\begin{tabular}{ p{0.15\linewidth}|p{0.20\linewidth} |p{0.10\linewidth}p{0.10\linewidth}p{0.08\linewidth}p{0.08\linewidth}p{0.08\linewidth}}
\hline
\multirow{5}{*}{{\textbf{2WikiMultiHop}}} &\textbf{Method} &  \textbf{EM} &  \textbf{delta EM}& \textbf{F1} &  \textbf{PR} & \textbf{RE} \\

 & \multirow{1}{*}{\small BM25+FS} 
  &  \small 27.8  &  \small &  \small 33.99  &  \small 33.01 &  \small 36.9  \\

& \multirow{1}{*}{\small GTR+FS} 
  &  \small 34.39  &  \small &  \small 43.7  &  \small 42.72 &  \small  48.55 \\

& \multirow{1}{*}{\small Cross-encoder+FS} 
  &  \small 33.1 &  \small  &  \small 39.4  &  \small 40.1 &  \small  47.54 \\

& \multirow{1}{*}{\small CoT-SC + VE} 
  &  \small 37.2 &  \small  &  \small-   &  \small - &  \small - \\

& \multirow{1}{*}{\small ReAct} 
  &  \small 28.0 &  \small  &  \small 38.5   &  \small -  &  \small -   \\

& \multirow{1}{*}{\small Self-Ask} 
  &  \small 37.3 &  \small  &  \small 48.8   &  \small -  &  \small -   \\

& \multirow{1}{*}{\small Iter-Retgen-6} 
  &  \small 35.5 &  \small  &  \small 48.1  &  \small -  &  \small -   \\

& \multirow{1}{*}{\textit{GenSco-stop+FS}} 
  &  \small 41.4 &  \small &  \small 51.23  &  \small 49.52 &  \small 56.09   \\

&\multirow{1}{*}{\textit{GenSco-max+FS}} 
  &  \small \textbf{43.2} &  \small 5.9 &\small \textbf{53.24}  &  \small \textbf{51.08} &  \small \textbf{58.28}   \\
\hline
\hline
\multirow{5}{*}{{\textbf{AdvHotPot}}}& \multirow{1}{*}{\small BM25+FS} 
  &  \small 46.75 &  \small &  \small 57.27  &  \small 58.23 &  \small 58.23 \\

& \multirow{1}{*}{\small GTR+FS} 
  &  \small 53.57 &  \small &  \small \textbf{63.36}   &  \small \textbf{64.29} &  \small 63.36   \\

& \multirow{1}{*}{\small Cross-encoder+FS} 
  &  \small 52.9 &  \small &  \small 60.1  &  \small 59.7 &  \small  59.4 \\

& \multirow{1}{*}{\small CoT-SC + VE} 
  &  \small \textbf{56.8} &  \small  &  \small  -  &  \small -  &  \small -   \\
& \multirow{1}{*}{\small \textit{GenSco-stop+FS}} 
  &  \small 55.52  &  \small -1.28&  \small 62.14  &  \small 63.08 &  \small  \textbf{64.03} \\

& \multirow{1}{*}{\small \textit{GenSco-max+FS}} 
  &  \small 54.87 &  \small &  \small 61.71  &  \small 62.65 &  \small 62.65   \\
\hline
\hline
\multirow{5}{*}{{\textbf{MuSiQue}}}& \multirow{1}{*}{\small BM25+FS} 
  &  \small 24.6 &  \small &  \small 29.2  &  \small 29.7 &  \small 31.2 \\

& \multirow{1}{*}{\small GTR+FS} 
  &  \small 25.4 &  \small &  \small 30.2   &  \small 34.7 &  \small 39.1   \\

& \multirow{1}{*}{\small Cross-encoder+FS} 
  &  \small 25.7 &  \small &  \small 28.2  &  \small 25.2 &  \small  34.1 \\

& \multirow{1}{*}{\small ReAct} 
  &  \small  23.4 &  \small  &  \small  37.0  &  \small -  &  \small -   \\

& \multirow{1}{*}{\small Self-Ask} 
  &  \small   27.6 &  \small &  \small  41.5  &  \small -  &  \small -   \\

& \multirow{1}{*}{\small Iter-Retgen-7} 
  &  \small  26.1  &  \small &  \small  42  &  \small -  &  \small -   \\

& \multirow{1}{*}{\small \textit{GenSco-stop+FS}} 
  &  \small 39.1  & &  \small \small 42.1  &  \small 39.7 &  \small  44.7 \\

& \multirow{1}{*}{\small \textit{GenSco-max+FS}} 
  &  \small \textbf{42.7} &  \small 15.1 &  \small \textbf{46.1}  &  \small \textbf{44.2} &  \small \textbf{47.9}   \\
\hline
\end{tabular}
\caption{Accuracy of various methods in terms of Exact Match (EM), F1-score (F1), Precision (PR) and Recall (RE). delta EM represents max EM of the two \textit{Tree} methods - max baseline EM. FS indicates few-shot prompting of the Generator LLM}
\label{main_table}
\end{table*}
In this section we will discuss the performance of our methodology on the three datasets and will compare with the baseline systems. Refer to the Table \ref{main_table} for the results.
\subsubsection{2WikiMultiHop}
Observe, performance across various correctness metrics, shows the improvements made by GenSco in comparison to all baseline systems. The jump in performance holds consistently true for most metrics, indicating the effectivenes of \textit{GenSco} for passage selection. Both our maximum hop based method (\textit{GenSco-max}) and the additional stopping-criteria based approach (\textit{GenSco-stop}) outperforms all baseline systems by a large margin validating the importance of sequential alignment between the retrieved passages and the question in the context of multi-hop Question Answering.

\subsubsection{Adversarial HotPot}
Here, we observe superior performance with significantly less variability across all baseline models. \textit{GenSco} does not outperform but is on par with the best-performing baselines, which can be attributed to the dataset's relatively smaller passage set. The advantages of our greedy search are not as pronounced because the provided passage set (for each question) is already small, thus limiting further gains with  filtering/selection.

\subsubsection{MuSiQue}

 With minimal potential reasoning shortcuts, this dataset is designed to necessitate connected reasoning of the model. Over this challenging dataset with harder distractor passages, our proposed method(s) achieve huge improvements across all four metrics over all the reported baseline systems. This is likely due to the presence of a sufficient number of candidate passages (up to 20) for each instance, allowing our algorithm to effectively explore a greedy selection. We observe that GenSco tends to excel when there are around 10 or more candidate passages on average per instance.
 
 In the rest of the experiments, we only use 2WikiMultiHop for further analysis.

\section{Discussion}
\subsection{What happens without the question decomposition?}
To gauge the role of question decomposition in GenSco, we setup an alternate implementation with the following two modifications to GenSco-max algorithm: 
\begin{enumerate}
    \item Instead of using the subquestion at the level to compute the log-likehood expression for passage selection in equation \ref{eq:score}, we use the original multi-hop question to compute the scores:
    \begin{equation*}
    \label{exp-4}
    \small
    \begin{multlined}
    score(n_{m,i}) =  nll_S(p(Q/P'[1:m-1], P[i])), \\ i\epsilon[1,k] \tag{4}
    \end{multlined}
    \end{equation*}
    \item For stopping criterion, we check if the best scoring paragraph at any level has already been selected along the greedy path. i.e. if
    $ P'[m] \quad \epsilon \quad P'[1:m-1]$.
\end{enumerate}
This variant is called GenSco-no-QD in Table~\ref{table:ablation} where we compare the correctness scores with two of the baseline methods and variants of GenSco on a subset of 250 instances from 2WikiMultiHop. The numbers severely drop down wrt to both GenSco-stop and GenSco-max but are still better than BM25. 

\begin{table}
\centering
\resizebox{\linewidth}{!}{%
\begin{tabular}{l|cccc}
\hline
\textbf{Method} &  \textbf{EM} & \textbf{F1} &  \textbf{PR} & \textbf{RE} \\
\hline
\multirow{1}{*}{\small BM25+FS} & \small 24.1&
    \small 29.76   & \small 28.8 & \small 32.64 \\
\hline
\multirow{1}{*}{\small GTR+FS} &
    \small  34.0 &\small 43.2 & \small 41.28 & \small 47.04\\
\hline
\multirow{1}{*}{\small GenSco-stop+FS} &
 \small 41.6  & \small 49.92& \small 48.96& \small 54.72 \\
\hline
\multirow{1}{*}{\small GenSco-max+FS} &
   \small 44.4  & \small 52.86 & \small 50.98& \small 57.58\\
\hline
\multirow{1}{*}{\small GenSco-no-QD+FS} &
   \small 31.97 & \small 38.4 & \small 37.44 & \small 41.28 \\

\hline

\end{tabular}%
}
\caption{Baseline and Variations of GenSco on 250 instances from 2WikiMultiHop (FS here stands for `FewShot')}
\label{table:ablation}
\end{table}

\subsection{Correctness vs Faithfulness}

\begin{table}
\centering
\resizebox{\linewidth}{!}{%
\begin{tabular}{l|cc}
\hline
\textbf{Method} &  \textbf{2WikiMultiHop} &  \textbf{AdvHotPot}\\
\hline
\multirow{1}{*}{\small BM25+FS} &
    \small  73.21  & \small86.98\\
\hline
\multirow{1}{*}{\small GTR+FS} &
    \small 83.6  &\small 88.05\\
\hline
\multirow{1}{*}{\small GenSco-stop+FS} &
 \small 75.55  & \small 75.37\\
\hline
\multirow{1}{*}{\small GenSco-max+FS} &
   \small 76  & \small 75.76\\
\hline
\end{tabular}%
}
\caption{Hallucination results (Metric: K-Precision)}
\label{hall_2Wiki}
\end{table}

\begin{figure}[h]
    \centering    \includegraphics[width=0.45\textwidth]{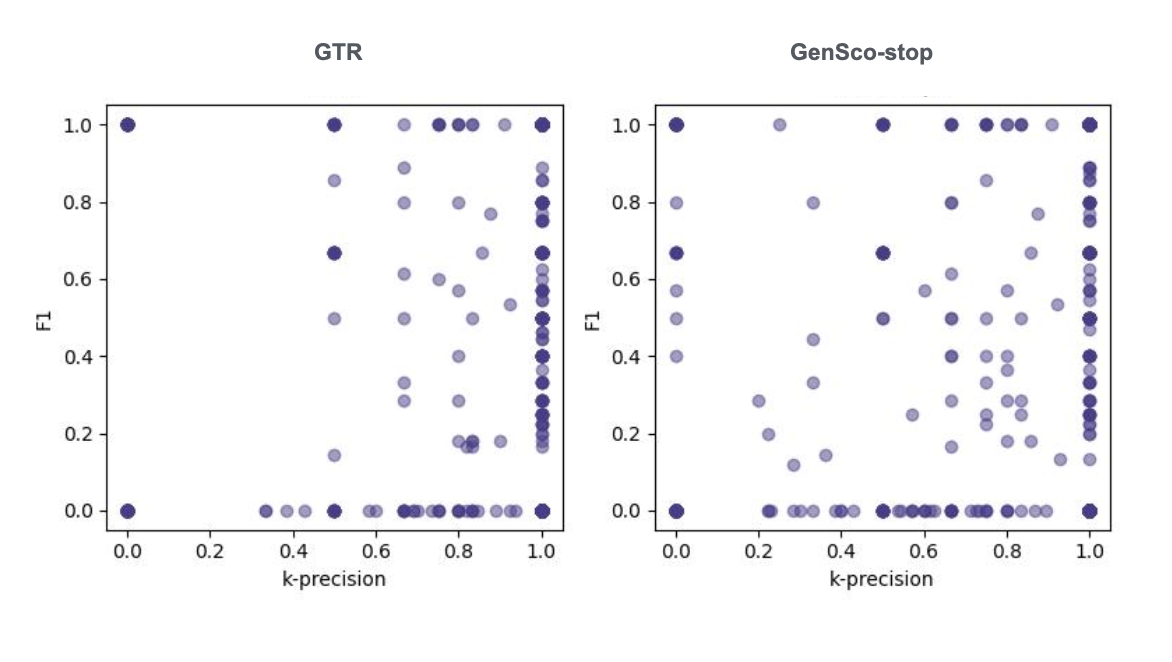}
\caption{Scatter plot of answers for 2WikiMultiHop}
   \label{fig:scatter}
\end{figure}

We also evaluated the model responses in comparison to baseline retrieval methods over faithfulness. K-precision~\cite{adlakha2023evaluating} serves to quantify how well a model's response is grounded within the provided passages. Consequently, if the generated answer shares more words with the passages, the k-precision value will be higher, regardless of its correctness. While GenSco methods are more accurate with respect to the ground truth answers, the GTR method achieves higher K-precision value (Table~\ref{hall_2Wiki}) for 2WikiMultiHop. 
Figure~\ref{fig:scatter} shows a scatter plot over a set of 960 instance responses for 2WikiMultihop from GenSco-stop and the GTR based method to observe the correlation between K-precision (indicating faithfulness) and F1 (which  is a  correctness metric). It's surprising that a higher K-precision score doesn't necessarily indicate increased correctness, as the data points are almost uniformly dispersed across the F1 axis for high k-precision values. The Pearson correlation coefficients for the two metrics for both GTR and GenSco-stop are hence low: $0.138$ and $0.238$, respectively. Typical RAG methods such as with GTR lack flexibility by relying on a fixed number of passages (k) for each instance typically setting k to be the estimated maximum number of hops required across the instances in the dataset which may result in higher value of k-precision. In contrast, our GenSco methods effectively tailor the number of passages for each multi-hop query(based on the decomposition), potentially leading to lower k-precision. However, the correctness measures, which are our primary concern, are consistently higher for our method indicating the importance of supplying only relevant passages to the generator LLM.

\begin{figure} [h]
    \centering    \includegraphics[width=0.45\textwidth]{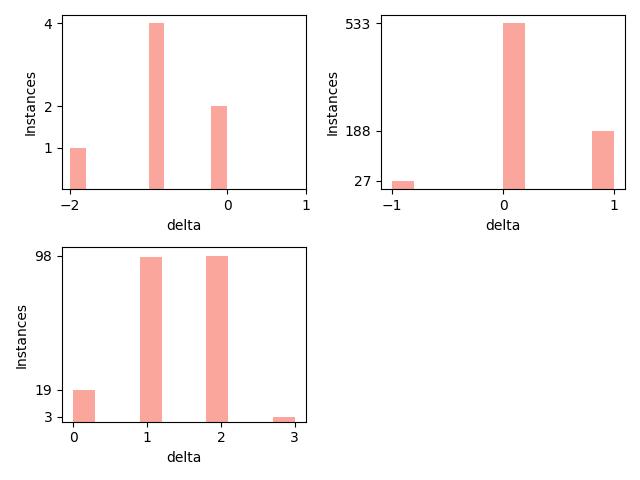}
\caption{Histogram of delta (number of supporting passages - number of passages retrieved by GenSco-stop)  for subsets of data with 1,2 and 4 supporting passages for 2WikiMultiHop dataset (left to right, top to bottom)}
   \label{fig:delta_hops}
\end{figure}
\begin{figure}[h]
    \centering    \includegraphics[width=0.45\textwidth]{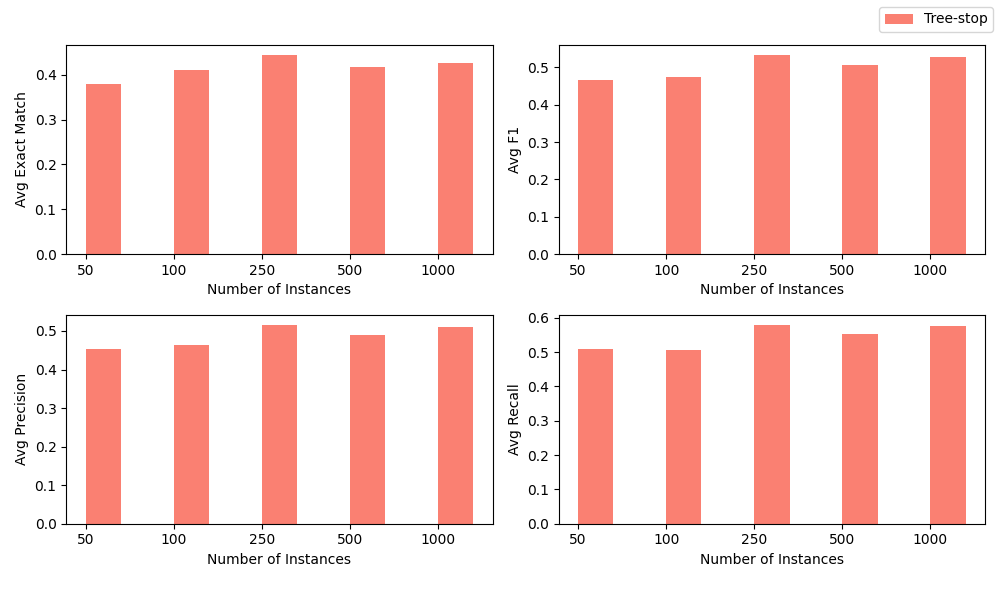}
\caption{Performance across different sized subsets of the 2WikiMultiHop dataset.}
   \label{fig:instances}
\end{figure}
 
\subsubsection{Retrieval Performance}

\begin{table}
\centering
\begin{tabular}{| p{0.3\linewidth}|p{0.08\linewidth} |p{0.20\linewidth}|p{0.15\linewidth}| }
\hline
\textbf{Method} &  \textbf{F1} &  \textbf{Precision}&  \textbf{Recall}\\
\hline
\multirow{1}{*}{\small GTR} &
    \small 56.2  & \small 42.46 & \small 88.57 \\
\hline
\multirow{1}{*}{\small GenSco-stop} &
    \small 83.32  & \small 93.51 & \small 78.35\\
\hline
\end{tabular}
\caption{Retrieval Performance of GTR and GenSco-stop on 2WikiMultiHop dataset}
\label{retr}
\end{table}

2WikiMultihop also provides details on which of the passages are supporting each question/answer (and which are just distractors). We use this information to contrast the retrieval performance of GenSco-stop with GTR. Table~\ref{retr} shows that GenSco-stop has a lower recall but much higher precision indicating that GenSco retrieves less irrelevant passages. 
Figure~\ref{fig:delta_hops} provides another rough idea on the difference in number of hops taken (number of sub-questions generated) by GenSco-stop vs the actual number of supporting passages specified in 2WikiMultiQA. The mod of delta as is seen in the plot mostly varies between 0 to 2 showing little divergence of GenSco from the ground truth.

\subsection{Does the order of the retrieved passages really matter?}
Since GenSco retrieves the passages in alignment with the implicit reasoning implied by the multi-hop question, the order in which these passages should be input to the Generator LLM is derived as part of our algorithm. To assess if the order of passages plays any role in the MHQA task, we performed a simple experiment where the retrieved passage sequences for GenSco-stop and GenSco-max are randomly shuffled before feeding into the Generator LLM. Table ~\ref{shuffle} shows how the correctness scores significantly degrade on randomizing the order. This experiment confirms the intuition based on which GenSco has been proposed as not just as a passage selection technique but also as a technique that identifies the passage \textit{sequence}. 
The retrieval baselines are not equipped to determine the order since the passage relevance is computed with respect to the  entire question as opposed to its specific parts/subquestions.

\subsection{Is the performance consistent over different sized subsets of the data?}
To check whether the performance remains consistent over different sizes of data subsets, we plot each of our four correctness metrics averaged over the instances in figure~\ref{fig:instances}. The aggregate values look fairly stable for GenSco-stop across various splits of 2WikiMultiHop dataset.

\begin{table}
\centering
\begin{tabular}{| p{0.4\linewidth}|p{0.08\linewidth} |p{0.08\linewidth}|p{0.08\linewidth}|p{0.08\linewidth}|}
\hline
\textbf{Method} &  \textbf{EM} & \textbf{F1} &  \textbf{PR} & \textbf{RE} \\
\hline
\multirow{1}{*}{\small GenSco-stop+FS } &
    \small  41.6   & \small 49.92 & \small 48.96 & \small 54.72\\
\hline
\multirow{1}{*}{\small GenSco-stop shuffled+FS} &
    \small 38.4  &\small 47.33 & \small 45.47 & \small 51.97\\
\hline
\multirow{1}{*}{\small GenSco-max+FS} &
 \small 44.4  & \small  52.86 & \small 50.98  & \small 57.58\\
\hline
\multirow{1}{*}{\small GenSco-max shuffled+FS} &
   \small 38  & \small 45.7  & \small 43.90 & \small 50.18\\
\hline
\end{tabular}
\caption{Scores with shuffling the selected passages before prompting the Generator for QA on 2WikiMultiHop (on 250 instances)
}
\label{shuffle}
\end{table}

\section{Conclusion}

We introduce an inference technique for passage sequence selection in multi-hop question-answering, outperforming baseline retrieval methods including recent SOTA systems on multiple datasets.
Experimental results demonstrate that the proposed approach not only captures pertinent passages but also offers a logical sequence for passages to be effectively processed in LLM prompts for multi-hop QA. Please note that our approach does not substitute the initial stage of passage retrieval from documents. Instead, it offers an finer-level filtering (and reordering) process over a set of already retrieved passages. Nonetheless, this approach can be used on top of mainstream retrieval systems for a more refined passage sequence selection ahead of LLM generation for MHQA.

\section{Limitations}
\begin{enumerate}
    \item For the generator LLM, we opted for GPT-3.5, a commercial LLM, as the free alternatives did not demonstrate comparable performance during our experiments. However, with the introduction of recent comparable open source models it is  possible that future research could investigate the use of free LLMs instead.
    \item Although our method is comparable with competitive baselines even on datasets with a small set of retrieved passages, we observe that it mostly thrives where we have an average of 10 or more candidate passages to choose from. However, this approach can complement traditional retrieval systems by offering a more precise selection of passage sequences before LLM generation for Multi-Hop Question Answering (MHQA).
\end{enumerate}

\bibliography{anthology,custom}

\newpage
\appendix
\section{Dataset} \label{app:dataset}
We have conducted experiments on three different datasets: \textbf{(i)} 2WikiMultiHop~\cite{ho2020constructing} offers a collection of $1000$ multi-hop question-answer pairs in the validation set beside $6$ pairs for few-shot learning. For each instance, there are $10$ passages provided, with some passages being pertinent to answering the question while others act as distractors. Along with that we also evaluated the technique on smaller contexts, hence worked on \textbf{(ii)} Adversarial HotPot~\cite{yang-etal-2018-hotpotqa,ye2022the}; it is smaller dataset comprising $308$ data instances, providing $4$ passages for every question-answer pair. For each instance, $2$ passages serve as supporting evidence, while the remaining 2 act as distractors.\textbf{(iii)} MuSiQue~\cite{trivedi-etal-2022-musique}: Following prior work ~\cite{shao2023enhancing,press2023measuring} , we evaluate on the 1252 questions from the Musique dev set categorized as 2-hop.\footnote{ The 3-hop and 4-hop questions are reported to be too intricate, even the authors of the paper found them challenging to comprehend at times~\cite{press2023measuring}.}

\section{Prompt templates}
\subsection{For Question Answering}
Table~\ref{table:qa_prompt} shows the structure of the template used for generating an answer using an instruction and few shots. The concatenated passage sequence along with the multihop question to be answered are then added.

\begin{table*}
\resizebox{\textwidth}{!}{%
\centering
\begin{tabular}{|p{0.07\textwidth}p{0.9\textwidth}|}
\hline
\multicolumn{2}{|l|}{Answer the question given the context. Here are a few examples:} \\

Question: & Which film was released earlier, Kistimaat or I'M Taraneh, 15? \\
Context: & Kistimaat is a 2014 Bangladeshi action film directed by Ashiqur Rahman and produced by Tiger Media Limited and The Abhi Pictures.The film features Arifin Shuvoo and Achol Akhe in lead roles while Misha Sawdagor plays the main antagonist in the film.The film is about a police officer and his fight against corruption.The film was released on Eid al- Adha, 6 October 2014, and was a commercial success.The movie was inspired by the 2010 Hindi film ``Dabangg".I'm Taraneh, 15 is a 2002 Iranian film directed by Rasul Sadrameli.The film was selected as the Iranian entry for the Best Foreign Language Film at the 75th Academy Awards, but it did not make the final shortlist. \\
Answer: & I'M Taraneh, 15 \\
\multicolumn{2}{|l|}{\textit{More shots can follow}}\\
\multicolumn{2}{|l|}{} \\
Question: & \textit{Question goes here} \\
Context: & \textit{Passages go here} \\
Answer: & \\
\hline
\end{tabular} %
}
\caption{Main prompt for drawing out the answer from the Generator}
\label{table:qa_prompt}
\end{table*}

\subsection{For Question Decomposition}
Table ~\ref{table:qd_prompt} shows the template used for drawing out the next subquestion from the Generator LLM conditioned on the history of subquestions and corresponding passages selected.

\begin{table*}
\resizebox{\textwidth}{!}{%
\begin{tabular}{|p{0.13\textwidth}p{0.60\textwidth}|}
\hline
\multicolumn{2}{|p{\linewidth}|}{I am going to give you a question. I want to decompose it into a series of subquestions. Each subquestion should be self-contained with all the information necessary to solve it.  Make sure not to decompose more than necessary or have any trivial subquestions. Do not repeat any subquestion. You’ll be evaluated on the simplicity, conciseness, and correctness of your decompositions. If no more subquestion could be drawn, please generate``\textless FIN\textgreater \textless/FIN\textgreater". Here are a couple of examples: }\\  
Question: & What are the other books from the author of ``The Good Earth"? \\
Subquestion 1: & Who is the author of the book ``The Good Earth"? \\
Subcontext 1: & The author of the book ``The Good Earth" is Pearl S. Buck. \\
Subquestion 2: & What are the titles of books written by Pearl S. Buck other than ``The Good Earth"? \\
\multicolumn{2}{|l|}{} \\
Question: & Which movie came out first ``Spiderman 2" or ``Batman Begins"?\\
Subquestion 1: & When was the release date of the movie ``Spiderman 2"? \\
Subcontext 1: & Spiderman 2 is a 2004 American superhero film based on the Marvel Comics character of the same name.\\
Subquestion 2: & When was the release date of the movie ``Batman Begins"? \\
\multicolumn{2}{|l|}{} \\
Question: & \textit{Question goes here} \\
Subquestion 1: & \textit{Subquestion 1 generated earlier goes here} \\
Subcontext 1: & \textit{Passage selected for Subquestion 1 goes here} \\
\multicolumn{2}{|l|}{\textit{k-2 pairs of follow up Subquestions and Subcontexts}}\\
Subquestion k: & \\
\hline
\end{tabular} %
}
\caption{Prompt for drawing out the next subquestion from the Generator.}
\label{table:qd_prompt}
\end{table*}

\subsection{For Stopping criterion}
Prompt for checking the stopping criterion (using Scorer LLM) is shown in Table~\ref{table:sc_prompt}. Note that this prompt is invoked both while conditioning on the decompostion upto and including subquestion k-1 and the decomposition until subquestion k individually. No shots are used here.

\begin{table*}
\resizebox{\textwidth}{!}{%
\begin{tabular}{|p{0.17\textwidth}p{0.8\textwidth}|}
\hline
\multicolumn{2}{|p{\textwidth}|}{Generate a question given its complete decomposition into subquestions along with the context containing answers to these subquestions.}\\
Context: & \textit{ Concatenation of Passages selected for Subquestions[1:k]} \\
Decomposition: & \textit{ Concatenation of Subquestions [1:k]} \\
Question: & \\
\hline
\end{tabular} %
}
\caption{Prompt for computing the value of the introduced stopping criterion using S. }
\label{table:sc_prompt}
\end{table*}

\subsection{For Passage selection}
Prompt for scoring (using Scorer) the passages based on the generated subquestion is shown in Table~\ref{table:ps_prompt}. Again, no shots are included (and the question generated by the scorer here is ignored). The scorer is prompted here only to draw out the likelihood scores from Scorer for the subquestion produced by the Generator LLM.
\begin{table*}
\resizebox{\textwidth}{!}{%
\begin{tabular}{|p{0.13\textwidth}p{0.84\textwidth}|}
\hline
\multicolumn{2}{|p{\textwidth}|}{Generate a question based on the context.}\\
Context: & \textit{The passage to be scored goes here} \\
Question: & \\
\hline
\end{tabular} %
}
\caption{Prompt for computing the relevance scores using S.}
\label{table:ps_prompt}
\end{table*}

\section{ Sample Trace of subquestions generated for GenSco-stop}
Table~\ref{table:trace} shows the 2-hop question logically processed by GenSco-stop. Relevance scores with respect to the subquestion are provided for each passage. The most relevant passage is selected for the first subquestion which feeds into generating the next subquestion. Similarly, the most relevant passage for the next subquestion is selected and subsequently the answer for the multi-hop question is generated.  GTR fails to retrieve one of the two relevant passages in top-5 and responds with an incorrect answer taken from a distractor passage.

\begin{table*}
\resizebox{\textwidth}{!}{%
\begin{tabular}{|p{0.05\textwidth}|p{0.70\textwidth}p{0.07\textwidth}|}
\hline
\multicolumn{3}{|c|}{} \\
\multicolumn{3}{|p{\textwidth}|}{What is the place of birth of the director of film The One And Only Ivan (Film)?}\\
\multicolumn{3}{|c|}{} \\
\hline
\multicolumn{3}{|c|}{} \\
\multicolumn{3}{|p{\textwidth}|}{\textbf{Who is the director of the film The One And Only Ivan (Film)?}} \\
\multicolumn{1}{|r|}{\small $1$} & Thea Sharrock (born 1976) is an English theatre and film director... & \multicolumn{1}{r|}{\small $-0.307$} \\
\multicolumn{1}{|r|}{\small $2$} &Peter Levin is an American director of film, television and theatre.  & \multicolumn{1}{r|}{\small $-0.334$} \\
\multicolumn{1}{|r|}{\small $3$} &The One and Only is a 1978 comedy film starring Henry Winkler, directed by Carl Reiner and written by Steve Gordon.  & \multicolumn{1}{r|}{\small $ -0.488$} \\
\multicolumn{1}{|r|}{\small $4$} &Andrei Virgil Ivan( born 4 January 1997) is a Romanian professional footballer who plays as a forward for Universitatea Craiova. & \multicolumn{1}{r|}{\small $-0.200$} \\
\multicolumn{1}{|r|}{\small $5$}&Katherine Alice Applegate( born October 9, 1956) is an American young adult... & \multicolumn{1}{r|}{\small $-0.654$} \\
\multicolumn{1}{|r|}{\small $6$} &Radu Ivan( born 17 July 1969) is a Romanian judoka who competed at three Olympic Games. & \multicolumn{1}{r|}{\small $-0.070$} \\
\multicolumn{1}{|r|}{\small $7$} &Ian Barry is an Australian director of film and TV. & \multicolumn{1}{r|}{\small $-0.230$} \\
\multicolumn{1}{|r|}{\small $8$} &\textbf{The One and Only Ivan is an upcoming American fantasy drama film directed by Thea Sharrock,...}& \multicolumn{1}{r|}{\small $-0.988$ }\\
\multicolumn{1}{|r|}{\small $9$} &Dávid Ivan( born 26 February 1995) is a Slovak professional footballer who plays as a midfielder for Serie B club Chievo.& \multicolumn{1}{r|}{\small $-0.296$ }\\
\multicolumn{1}{|r|}{\small $10$}&Marian Ivan( born 1 June 1969 in Bucharest) is a retired Romanian footballer... & \multicolumn{1}{r|}{\small $-0.228$ }\\
\hline
\multicolumn{3}{|c|}{} \\
\multicolumn{3}{|p{\textwidth}|}{\textbf{What is the place of birth of Thea Sharrock? }}\\
\multicolumn{1}{|r|}{\small $1$} &\textbf{Thea Sharrock (born 1976) is an English theatre and film director...} & \multicolumn{1}{r|}{\small $-1.617$}\\
\multicolumn{1}{|r|}{\small $2$} &Peter Levin is an American director of film, television and theatre.  & \multicolumn{1}{r|}{\small $-0.101$} \\
\multicolumn{1}{|r|}{\small $3$} &The One and Only is a 1978 comedy film starring Henry Winkler, directed by Carl Reiner and written by Steve Gordon.  & \multicolumn{1}{r|}{\small $  -0.594$} \\
\multicolumn{1}{|r|}{\small $4$}&Andrei Virgil Ivan( born 4 January 1997) is a Romanian professional footballer who plays as a forward for Universitatea Craiova. & \multicolumn{1}{r|}{\small $-0.213$} \\
\multicolumn{1}{|r|}{\small $5$} &Katherine Alice Applegate( born October 9, 1956) is an American young adult... & \multicolumn{1}{r|}{\small $-0.481$} \\
\multicolumn{1}{|r|}{\small $6$} &Radu Ivan( born 17 July 1969) is a Romanian judoka who competed at three Olympic Games. & \multicolumn{1}{r|}{\small $ -0.207$} \\
\multicolumn{1}{|r|}{\small $7$} &Ian Barry is an Australian director of film and TV. & \multicolumn{1}{r|}{\small $-0.104$} \\
\multicolumn{1}{|r|}{\small $8$} &The One and Only Ivan is an upcoming American fantasy drama film directed by Thea Sharrock,...& \multicolumn{1}{r|}{\small $-1.164$ }\\
\multicolumn{1}{|r|}{\small $9$}&Dávid Ivan( born 26 February 1995) is a Slovak professional footballer who plays as a midfielder for Serie B club Chievo.& \multicolumn{1}{r|}{\small $-0.347$ }\\
\multicolumn{1}{|r|}{\small $10$} &Marian Ivan( born 1 June 1969 in Bucharest) is a retired Romanian footballer... & \multicolumn{1}{r|}{\small $-0.328$ }\\
\hline
\multicolumn{3}{|c|}{} \\
\multicolumn{3}{|p{\textwidth}|}{\textbf{Answer (with the passages(two passages in bold) above selected using GenSco-stop)}: London, England  }\\ 
\multicolumn{3}{|c|}{} \\
\hline
\hline
& \multicolumn{2}{|c|}{} \\
\multicolumn{1}{|r|}{\small $1$} & \multicolumn{2}{|l|}{The One and Only Ivan is an upcoming American fantasy drama film directed by Thea Sharrock...} \\
\multicolumn{1}{|r|}{\small $2$} & \multicolumn{2}{|l|}{Marian Ivan( born 1 June 1969 in Bucharest) is a retired Romanian footballer......} \\
\multicolumn{1}{|r|}{\small $3$} & \multicolumn{2}{|l|}{Dávid Ivan( born 26 February 1995) is a Slovak professional footballer who plays as a midfielder for Serie B club Chievo.}\\ 
\multicolumn{1}{|r|}{\small $4$} & \multicolumn{2}{|l|}{Andrei Virgil Ivan( born 4 January 1997) is a Romanian professional footballer who plays as a forward for Universitatea Craiova. } \\
\multicolumn{1}{|r|}{\small $5$} & \multicolumn{2}{|l|}{Radu Ivan( born 17 July 1969) is a Romanian judoka who competed at three Olympic Games. } \\
\hline
\multicolumn{3}{|c|}{} \\
\multicolumn{3}{|p{\textwidth}|}{\textbf{Answer (with passages above selected using GTR)}: Bucharest  }\\ 
\multicolumn{3}{|c|}{} \\
\hline
\end{tabular} %
}
\caption{Trace of an example from 2WikiMultiHop using GenSco-stop method and GTR. The generated subquestions and selected passages are bolded for GenSco-stop. Passages are truncated for space limits. }
\label{table:trace}
\end{table*}
\section{Computational cost}

Given that we use GPT-3.5 as the Generator, our computational efficiency is constrained by the maximum number of API calls allowed by OpenAI within a given time frame. For each query, we perform up to $O(k)$ inference operations (where $k$ represents the number of passages) using LLAMA-2 and GPT-3.5, and finally, we call GPT-3.5 once to generate an answer to the multi-hop question. Replacing GPT-3.5 with a competitive open-source large language model (LLM) for generation in GenSco could potentially reduce the turnaround time and warrants further exploration in future work.

\begin{table}
\centering
\resizebox{\linewidth}{!}{%
\begin{tabular}{l|cccc}
\hline
\textbf{Method} &  \textbf{EM} & \textbf{F1} &  \textbf{PR} & \textbf{RE} \\
\hline
\multirow{1}{*}{\small GenSco-stop+FS} &
 \small 41.6  & \small 49.92& \small 48.96& \small 54.72 \\
\hline
\multirow{1}{*}{\small GenSco-max+FS} &
   \small 44.4  & \small 52.86 & \small 50.98& \small 57.58\\
\hline
\multirow{1}{*}{\small GenSco-stop+FS temp 0.5} &
   \small 41.60  & \small 50.67 & \small 48.76 & \small 55.45\\
\hline
\multirow{1}{*}{\small GenSco-max+FS temp 0.5} &
   \small 45.20  & \small 53.54 & \small 51.62 & \small 58.32 \\
\hline
\end{tabular}%
}
\caption{Variations of GenSco on 250 instances from 2WikiMultiHop (FS here stands for `FewShot') with different temperatures}
\label{table:ablation-temp}
\end{table}
\section{Temperature}
Setting  the temperature (for Generator) to a value $0.5$ (see Table~\ref{table:ablation-temp}) gives small but consistent gains across the correctness metric. 
With a temperature $<1$, the model becomes more deterministic, leading to more focused responses which suits QA task at hand. Note that the default temperature is set to $0$ wherever not mentioned otherwise.

\section{Can just longer context solve the problem instead?}
Even though the language models can now take upto and more than 128k tokens \cite{DBLP:journals/corr/abs-2402-13753} as context into the prompts, it does not solve the problem of \textit{reasoning} over long documents. As of now, multihop reasoning/QA remains a challenge even for the SOTA LLMs\cite{mavi2022survey}. Effectively aligning the correct passages (in the prompt) remains a critical challenge that GenSco can address to prevent hallucinations due to distracting passages in the prompt. Therefore, we posit that GenSco can serve as a fundamental solution for the LLMs, regardless of their maximum allowed context lengths.

\section{GenSco is an inference method, hence not directly comparable to the full finetuning setup}
With respect to the leaderboard for the datasets, GenSco scores are not directly comparable since GenSco is only an inference method that does not rely on the respective training subsets for these datasets. Fine-tuning Large Language Models (LLMs) in different settings, especially the commercial LLMs can be challenging and sometimes impractical. We draw inspiration from different approaches researchers have tried to make zero-shot/few-shot settings effective for the MHQA~\cite{gao2024retrievalaugmented}. With such approaches, including our work, passage selection is done by leveraging the LLMs with in-context learning for Multihop Question Answering (MHQA). The SOTA models for 2WikiMultiHop in the leaderboard and similarly for other reported datasets, are trained on the respective train sets while we propose an inference method assuming no access to the training examples. This is the reason why the best numbers on the dataset leaderboards are not directly comparable against our setting, but comparable to other inference based approaches in prior work.
\end{document}